\documentclass[nonacm]{acmart}
\AtBeginDocument{%
  }

\usepackage{lipsum}
\usepackage{soul}
\usepackage{algorithm}
\usepackage{algorithmic}

\newcommand\blfootnote[1]{%
  \begingroup
  \renewcommand\thefootnote{}\footnote{#1}%
  \addtocounter{footnote}{-1}%
  \endgroup
}

\begin{document}

\title{SRMU: Relevance-Gated Updates for Streaming Hyperdimensional Memories}

\author{Shay Snyder}
\orcid{0000-0002-3369-3478}
\affiliation{%
  \institution{George Mason University}
  \city{Fairfax}
  \state{Virginia}
  \country{USA}
}
\email{ssnyde9@gmu.edu}

\author{Andrew Capodieci}
\orcid{0009-0007-1272-6465}
\affiliation{%
  \institution{Neya Systems}
  \city{Pittsburgh}
  \state{Pennsylvania}
  \country{USA}}
\email{acapodieci@neyarobotics.com}

\author{David Gorsich}
\orcid{0000-0003-2961-1393}
\affiliation{%
  \institution{U.S. Army Ground Systems}
  \city{Warren}
  \state{Michigan}
  \country{USA}
}
\email{david.j.gorsich.civ@army.mil}

\author{Maryam Parsa}
\orcid{0000-0002-4855-4593}
\affiliation{%
 \institution{George Mason University}
 \city{Fairfax}
 \state{Virginia}
 \country{USA}}
 \email{mparsa@gmu.edu}

\renewcommand{\shortauthors}{Snyder et al.}

\begin{abstract}
Sequential associative memories (SAMs) are difficult to build and maintain in real-world streaming environments, where observations arrive incrementally over time, have imbalanced sampling, and non-stationary temporal dynamics.
Vector Symbolic Architectures (VSAs) provide a biologically-inspired framework for building SAMs.
Entities and attributes are encoded as quasi-orthogonal hyperdimensional vectors and processed with well defined algebraic operations.
Despite this rich framework, most VSA systems rely on simple additive updates, where repeated observations reinforce existing information even when no new information is introduced.
In non-stationary environments, this leads to the persistence of stale information after the underlying system changes.
In this work, we introduce the Sequential Relevance Memory Unit (SRMU), a domain- and cleanup-agnostic update rule for VSA-based SAMs.
The SRMU combines temporal decay with a relevance gating mechanism.
Unlike prior approaches that solely rely on cleanup, the SRMU regulates memory formation by filtering redundant, conflicting, and stale information before storage.
We evaluate the SRMU on streaming state-tracking tasks that isolate non-uniform sampling and non-stationary temporal dynamics.
Our results show that the SRMU increases memory similarity by $12.6\%$ and reduces cumulative memory magnitude by $53.5\%$. 
This shows that the SRMU produces more stable memory growth and stronger alignment with the ground-truth state.
\end{abstract}





\maketitle

\section{Introduction}
\label{sec:introduction}

\blfootnote{DISTRIBUTION STATEMENT A. Approved for public release; distribution is unlimited. OPSEC \#10535.}

Sequential associative memories (SAMs) serve an important role in streaming perception and monitoring systems, where information must be integrated as observations arrive over time~\cite{kleyko2023part2}.
Examples include distributed sensing~\cite{chen2025federated}, long-running robotic deployments~\cite{snyder2026brain}, and other partially observed environments where only a subset of the system is visible at any given time~\cite{snyder2025generalizable}.
In these scenarios, the memory must do more than simply accumulate observations.
It must preserve useful information about previously observed entities while remaining responsive to new evidence and changes in the underlying system.
This challenge becomes especially difficult when observations are unevenly sampled and when the latent state evolves over time.

A SAM is a memory that incrementally stores associations between entities and their attributes, states, or context as new observations arrive~\cite{kleyko2023part2}.
Vector Symbolic Architectures (VSAs) provide a structured and biologically inspired framework for implementing SAMs~\cite{kleyko2023part2}.
In a VSA-based SAM, keys and values are represented as quasi-orthogonal hyperdimensional vectors.
Keys and values are combined through a binding operation to form an associative representation, and multiple associations are accumulated through bundling~\cite{Kleyko_2022}.
To recover stored information, a query key is unbound from memory and the resulting estimate is compared against candidate values using cleanup and similarity operations~\cite{Kleyko_2022}.
These operations make VSAs an attractive framework SAMs because they support compact distributed representations, incremental updates, and similarity-based retrieval.

However, most VSA-based SAMs rely on simple additive updates, where new observations are superposed directly into memory~\cite{kleyko2023part2}.
This update rule treats all observations as equally informative, regardless of whether the memory already contains the same information or the underlying association has changed.
As a result, repeated observations can disproportionately reinforce existing content, while outdated information may persist after environment changes.
These effects become especially problematic in streaming settings with non-uniform sampling and non-stationary temporal dynamics, where the memory must balance persistence and adaptability~\cite{snyder2025generalizable, snyder2026brain}.

Several strategies have been explored to address these limitations, including temporal decay mechanisms that attenuate older information~\cite{frady2018sequence} and retrieval-stage cleanup or post-processing methods that improve robustness in specific domains~\cite{Renner2024, snyder2026brain}.
While these approaches are effective, they do not regulate how each new observation should influence the memory based on its relationship to the current memory state.
In particular, they do not explicitly distinguish between redundant, novel, and conflicting information during memory formation.

In this work, we introduce the Sequential Relevance Memory Unit (SRMU), a domain-agnostic update rule for VSA-based SAMs. 
The SRMU combines temporal decay with a relevance gating mechanism that modulates how strongly new observations are incorporated into memory.
Rather than treating all observations uniformly, the SRMU suppresses redundant updates while remaining responsive to new information.
This enables the memory to remain more stable under non-uniform sampling and rapidly adapt in non-stationary environments.

We evaluate the SRMU on synthetic state-tracking tasks designed to isolate key challenges in SAMs.
We compare the SRMU against naive additive and temporal decay memory baselines.
Our results show the SRMU's information-aware memory reduces the cumulative memory magnitude by $53.5\%$ and improves ground truth alignment by $12.6\%$.

The major contributions of this work are summarized as follows:
\begin{itemize}
    \item We introduce the Sequential Relevance Memory Unit (SRMU), a domain-agnostic update rule that regulates memory formation based on the relationship between new observations and the current memory state.
    \item We compare the SRMU against naive additive and temporal decay baselines using metrics that capture retrieval fidelity and memory stability.
    \item We show that the SRMU reduces memory growth by $53.5\%$ and improves retrieval similarity by $12.6\%$.
\end{itemize}
\section{Problem Statement}

Many systems sequentially integrate information as observations arrive over time~\cite{snyder2026brain}.
Examples include robotics~\cite{Renner2024}, reinforcement learning~\cite{snyder2025generalizable}, and cognitive architectures~\cite{karunaratne2024rolenoisefactorizersdisentangling}.
A common abstraction for this problem is the formation of sequential associative memories (SAMs), where observations are represented as key–value pairs that are incrementally integrated into a shared memory~\cite{kleyko2023part2}.
A \textit{key} identifies the context associated with an observation, while the \textit{value} represents the desired information.


The goal is to maintain a memory representation that incrementally integrates observations while preserving the ability to retrieve information associated with individual keys.
However, sequential observations may be redundant, conflicting, or subject to nonstationary temporal dynamics, making it unclear how each new observation should influence the existing memory~\cite{frady2018sequence}.
This raises a central question: how should memory updates be regulated so that informative observations are incorporated while redundant or stale information is suppressed?

\section{VSA-Based Sequential Memories}
\label{section:vsa-seq-mem}

A natural approach to implementing SAMs is through hyperdimensional computing (HDC) and vector symbolic architectures (VSAs)~\cite{Kleyko_2022}.
These frameworks provide a structured mechanism for representing key–value associations in high-dimensional spaces~\cite{kanerva1994spatter}.
VSAs represent symbols and structured relationships using high-dimensional vectors and algebraic operations such as \emph{binding} and \emph{bundling}~\cite{plate2003holographic}.
Binding ($\otimes$) combines keys and values into composite representations, while bundling ($+$) aggregates multiple associations into a shared memory vector~\cite{plate1991holographic}.
These operations enable simple, highly parallel updates and support efficient similarity-based retrieval~\cite{plate2003holographic}.
These operations have also been proposed as a biologically plausible form of associative memory~\cite{frady2020resonator}.

In a VSA memory, a key–value observation is encoded as
\begin{equation}
f_t = k_t \otimes v_t,
\end{equation}
where $\otimes$ denotes a binding operation between the key $(k_t\in\mathcal{K})$ and value $(v_t\in\mathcal{V})$ vectors. Sequential observations are then integrated into memory through bundling,
\begin{equation}
M_t = M_{t-1} + f_t.
\end{equation}
This allows multiple associations to be accumulated within a single memory vector while supporting approximate retrieval through inverse binding and similarity operations.

However, bundling assumes that all observations contribute equally to the memory representation~\cite{Kleyko_2022}.
In streaming environments, this assumption often breaks down, where observations may be redundant, unevenly sampled, or generated by systems that evolve over time~\cite{snyder2025generalizable, snyder2026brain}.
Therefore, memories can degrade through over-reinforcement, interference, and stale information retention. 
In the next section, we examine these failure modes in detail.

\section{Failure Modes of Streaming VSA Memories}
\label{section:failure-modes}

VSA-based SAMs operate by incrementally incorporating observations into a shared memory through bundling~\cite{kleyko2023part2}. 
While this update mechanism enables efficient online learning, it implicitly assumes that all observations are equally informative. 
In real-world streaming environments, this assumption often fails~\cite{snyder2026brain}.

Two failure modes are particularly common: \emph{sampling imbalance} and \emph{non-stationary dynamics}. 
Both arise from properties of the underlying process and interact directly with the uniform bundling operation used in most VSA memories~\cite{kleyko2023part2, furlong2024modelling, snyder2026brain}.

\textit{Sampling Imbalance: }
In many systems, some keys are observed more frequently than others~\cite{snyder2026brain}. 
As a result, frequent observations are repeatedly reinforced through repeated updates, while rarely observed keys receive significantly less. 
This leads to memory representations that are dominated by frequently sampled associations and making it difficult to recover information associated with sparse observations~\cite{snyder2026brain}.
Importantly, repeated observations often do not introduce new information. 
With the standard bundling operation, redundant updates continue to increase the magnitude of already well-represented associations, leading to an inefficient use of representational capacity~\cite{snyder2025generalizable}.

\textit{Non-Stationary Associations: }
In many environments, the relationships between keys and values evolve over time due to gradual drift or abrupt state changes in the underlying process~\cite{dumont2025symbols}. 
SAMs must therefore balance the persistence of previously stored information with the ability to rapidly incorporate new evidence. 
A common approach is to introduce temporal decay so that older contributions are gradually attenuated~\cite{frady2018sequence}. 
While decay improves adaptability, it operates uniformly across the memory and does not distinguish between stale and relevant information. 
As a result, decay alone may either retain outdated information for too long or prematurely remove relevant information.


\textit{Implications for Streaming Memory Updates: }
These challenges highlight a fundamental limitation of simple bundling and time-decayed VSA memories: the update rule does not account for the relationship between new observations and the current memory state. 
Redundant observations are over-reinforced, while meaningful changes may not be efficiently incorporated in non-stationary environments.
This suggests that VSA-based SAMs require update mechanisms that modulate new observations based on their information content. 
In particular, updates should suppress redundant information while remaining responsive to changes under non-stationary dynamics.

Input-level filtering provides one possible mitigation strategy by suppressing redundant observations prior to encoding. 
However, such approaches typically rely on domain-specific assumptions about the environment~\cite{gallego2020event, duong2022autonomous}, limiting their generality across tasks and observation modalities. 

Alternatively, prior work has addressed these challenges at the retrieval stage through associative cleanup mechanisms, including resonator networks~\cite{frady2020resonator, Renner2024} and modern Hopfield networks~\cite{ ramsauer2020hopfield}. 
While these methods can improve readout fidelity, they operate only during retrieval and do not influence how information is written into memory. 
As a result, interference introduced during streaming updates remains embedded in the representation.

In contrast, we seek a representation-level update mechanism that operates directly on encoded key–value associations. 
By modulating how new information is incorporated into memory, this approach provides a domain-agnostic solution that generalizes across VSA-based associative memory systems.

To address these challenges, we introduce the \emph{Sequential Relevance Memory Unit (SRMU)}, a VSA-based update mechanism that conditions each update on the relationship between new observations and the current memory state. 
Rather than treating all observations uniformly, SRMU modulates updates based on their information content, suppressing redundant observations while emphasizing new information. 

\begin{figure}
    \centering
    \includegraphics[width=0.98\linewidth]{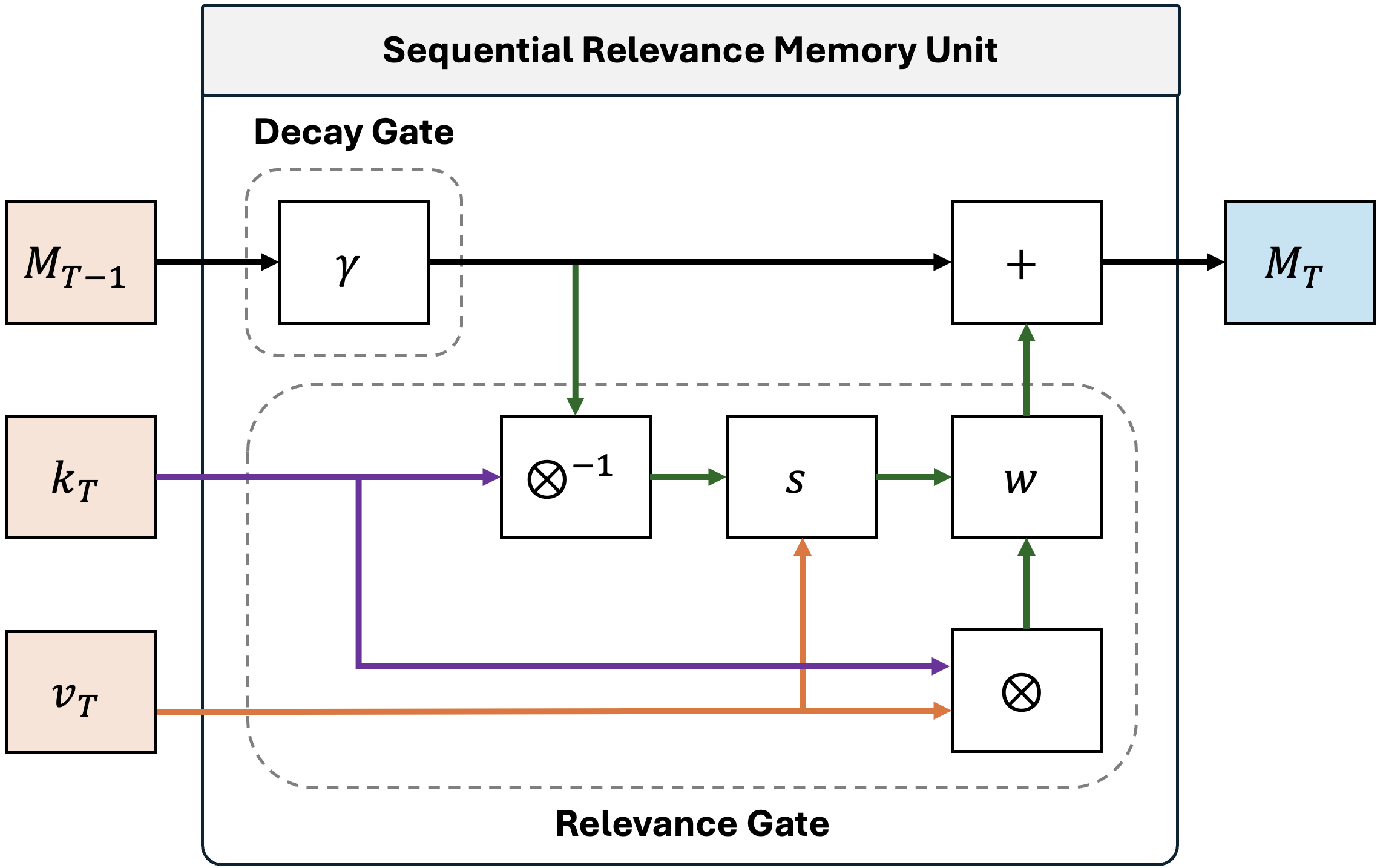}
    \caption{The Sequential Relevance Memory Unit}
    \Description{The Sequential Relevance Memory Unit}
    \label{fig:srmu}
\end{figure}

\begin{algorithm}
\caption{Sequential Relevance Memory Unit (SRMU) Update}
\label{alg:srmu}
\begin{algorithmic}
\REQUIRE Previous memory state $M_{t-1} \in \mathbb{C}^D$
\REQUIRE Key hypervector $k_t \in \mathbb{C}^D$
\REQUIRE Value hypervector $v_t \in \mathbb{C}^D$
\REQUIRE Temporal persistence parameter $\gamma \in (0,1]$
\ENSURE Updated memory state $M_t \in \mathbb{C}^D$

\STATE $\widetilde{M}_{t-1} \gets \gamma M_{t-1}$

\STATE $\hat{v}_t \gets \widetilde{M}_{t-1} \otimes^{-1} k_t$

\IF{$\|\hat{v}_t\|_2 > 0$ and $\|v_t\|_2 > 0$}
    \STATE $s \gets \dfrac{\left|\langle \hat{v}_t, v_t \rangle\right|}{\|\hat{v}_t\|_2 \|v_t\|_2}$
\ELSE
    \STATE $s \gets 0$
\ENDIF

\STATE $w \gets 1 - s$
\STATE $\Delta M \gets w (k_t \otimes v_t)$
\STATE $M_t \gets \widetilde{M}_{t-1} + \Delta M$

\STATE \RETURN $M_t$
\end{algorithmic}
\end{algorithm}

\section{Sequential Relevance Memory Unit (SRMU)}
\label{section:srmu}

The failure modes described in Section~\ref{section:failure-modes} highlight that VSA-based SAMs should distinguish between redundant observations and meaningful changes. 
In particular, updates should suppress repeated information while remaining responsive to new information.
To address these challenges, we introduce the \emph{Sequential Relevance Memory Unit (SRMU)}, an information-aware update mechanism that regulates how observations are incorporated into VSA-based SAMs.

The SRMU is defined in terms of generic VSA operations, including binding, unbinding, and similarity, and is therefore applicable across a broad class of VSA representations~\cite{Kleyko_2022}. 
For concreteness, we present the formulation using Fourier Holographic Reduced Representations (FHRR)~\cite{plate2003holographic}.

The SRMU exploits the key–value structure of associative memories. 
Given a key hypervector $k_t \in \mathbb{C}^D$ and value hypervector $v_t \in \mathbb{C}^D$ representing the observation at time $t$, the memory state $M_{t-1} \in \mathbb{C}^D$ is updated as

\begin{equation}
M_t = \operatorname{SRMU}(M_{t-1}, k_t, v_t; \gamma),
\end{equation}

where $\gamma \in (0,1]$ is a temporal decay parameter. For reference, the standard superposition update for key–value VSA memories is
\begin{equation}
M_t = M_{t-1} + k_t \otimes v_t ,
\end{equation}
where $\otimes$ denotes binding and $+$ denotes bundling~\cite{Kleyko_2022}. 
This update treats every observation as entirely new information, regardless of whether the memory already contains similar information.
The SRMU modifies this rule through two complementary mechanisms: temporal decay for handling non-stationary dynamics, and relevance-gating for suppressing redundant information.

\subsection{Temporal Decay}

SAMs should be responsive to more recent information when the underlying system changes over time. 
At the same time, they cannot forget too aggressively, especially when observations are sparse or unevenly sampled. 
This creates a trade-off between preserving useful context and adapting to new information.

A common way to handle this is through temporal decay, where older information is attenuated over time~\cite{frady2018sequence}. 
In the SRMU, this produces a decayed prior memory,
\begin{equation}
    \widetilde{M}_{t-1} = \gamma M_{t-1},
\end{equation}
where $0 < \gamma \leq 1$. 
Smaller values of $\gamma$ place more emphasis on recent observations, while larger values preserve more temporal context.

Temporal decay reduces the influence of outdated information, but it does so uniformly across the memory. 
It does not distinguish between information that is stale and information that is still relevant but has not been observed recently. 
As a result, decay alone can create a trade-off: if $\gamma$ is too large, outdated information may remain in memory for too long; if $\gamma$ is too small, useful information may be attenuated
prematurely.

In the SRMU, the decayed memory $\widetilde{M}_{t-1}$ is not only used to retain prior state, but also to provide a reference for the update. 
This allows the memory to compare a new observation against what is already stored and determine whether that observation is redundant or reflects a meaningful change. 
This comparison forms the basis of the relevance-gated update described next.

\subsection{Key-Conditioned Relevance}

To determine whether an observation introduces new information, the SRMU first retrieves the value currently associated with the observed key. 
In VSA-based SAMs this is accomplished by unbinding the key from the prior memory,
\begin{equation}
\hat{v}_t = \widetilde{M}_{t-1} \otimes^{-1} k_t ,
\end{equation}
where $\otimes^{-1}$ denotes the unbinding operation~\cite{Kleyko_2022}. 
For FHRR representations this corresponds to elementwise multiplication with the complex conjugate of the key vector~\cite{frady2022computing}.
The retrieved vector $\hat{v}_t$ represents the value currently associated with key $k_t$.

\subsection{Relevance-Gated Updates}

SRMU regulates the magnitude of the memory update based on the similarity between the retrieved value $\hat{v}_t$ and the newly observed value $v_t$. 
Similarity is measured using cosine similarity,

\begin{equation}
s(\hat{v}_t, v_t) =
\frac{|\langle \hat{v}_t, v_t \rangle|}
{\|\hat{v}_t\|_2 \|v_t\|_2},
\end{equation}

where $\langle \cdot,\cdot \rangle$ denotes the Hermitian inner product.

A novelty weight is then defined as

\begin{equation}
w = 1 - s(\hat{v}_t, v_t).
\end{equation}

This weight suppresses updates when the retrieved value already matches the observation and increases when the observation differs from the current memory. 

\subsection{Final SRMU Update Rule}

Combining temporal persistence with relevance gating yields the final SRMU update rule

\begin{equation}
M_t = \widetilde{M}_{t-1} + w (k_t \otimes v_t),
\end{equation}

where

\begin{equation}
\widetilde{M}_{t-1} = \gamma M_{t-1}
\end{equation}

and

\begin{equation}
w = 1 - s(\hat{v}_t, v_t), \qquad
\hat{v}_t = \widetilde{M}_{t-1} \otimes^{-1} k_t.
\end{equation}

By conditioning updates on the retrieved value associated with the current key, SRMU regulates how new observations are incorporated into the memory. 
Redundant observations are suppressed, preventing disproportionate accumulation from non-uniform sampling, while deviations from the current estimate produce stronger updates that enable rapid adaptation in non-stationary environments.

\section{Results}

\begin{figure}
    \centering
    \includegraphics[width=0.9\linewidth]{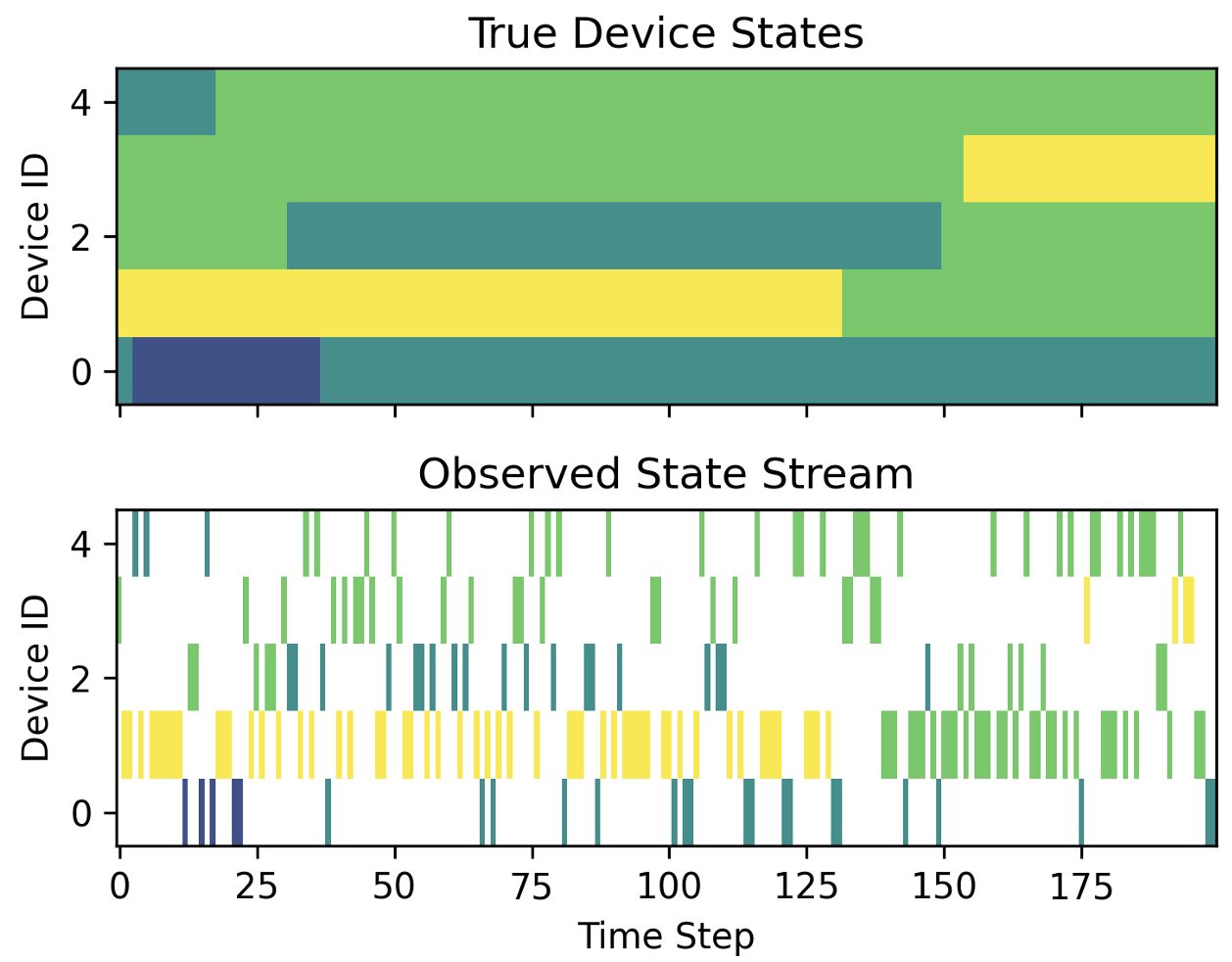}
    \caption{Example device state space and observation stream}
    \Description{Example device state space and observation stream}
    \label{fig:exp-example}
\end{figure}

We evaluate the Sequential Relevance Memory Unit (SRMU) in a synthetic device-health monitoring problem designed to isolate common failure modes of SAMs.
In this problem $K=5$ devices occupy one of $S=5$ ordinal health states with the latent state of each device evolving at every timestep.
The task is therefore not simply to memorize incoming observations, but to maintain an accurate estimate of the full system state under partial, noisy and potentially biased observations.

For each timestep, the hidden state of every device is updated according to a stochastic transition process.
Local changes are modeled as small $\pm1$ ordinal drifts with probability $p_\text{drift}$, while abrupt transitions are modeled as jumps to a random state with probability $p_\text{jump}$.
After the latent state is updated, one device is selected for observation according to a uniform or non-uniform sampling schedule.
All observations are subject to a 5\% change of noise where the observation will be $\pm1$ from the true value.
An example ground truth device state space and observation stream is shown in Figure~\ref{fig:exp-example}.

We report two complementary metrics. First, the cosine similarity measures the alignment between the learned memory and ground-truth memory. Second, the norm of the memory vector is tracked to determine whether performance is being achieved through stable representations or uncontrolled memory growth. Together, these metrics distinguish retrieval fidelity, memory stability, and representation efficiency.
All experiments were performed with a vector dimensionality of $256$.

\begin{table*}[]
\caption{Results for each memory model across all experiments.}
\label{table:exp-results}
\begin{tabular}{c|cc|cc|cc}
\hline
& \multicolumn{2}{c|}{Experiment 1} & \multicolumn{2}{c|}{Experiment 2} & \multicolumn{2}{c}{Experiment 3}            \\ \hline
\textbf{Memory Model}    & Cosine Sim     & Magnitude      & Cosine Sim     & Magnitude      & Cosine Sim     & Magnitude      \\ 
\hline
Naive                    & 0.832          & 1650.79        & 0.492          & 8115.55        & 0.397          & 9337.07        \\
SRMU ($\gamma=1.0$)      & \textbf{0.942} & 726.40         & 0.515          & 5665.24        & 0.455          & 5670.31        \\
Temporal ($\gamma=0.95$) & 0.817          & 168.93         & 0.884          & 136.65         & 0.755          & 159.26         \\
SRMU ($\gamma=0.95$)     & 0.909          & \textbf{75.96} & \textbf{0.907} & \textbf{79.06} & \textbf{0.850} & \textbf{74.01} \\ 
\hline
\end{tabular}
\end{table*}

\begin{figure}
    \centering
    \includegraphics[width=0.9\linewidth]{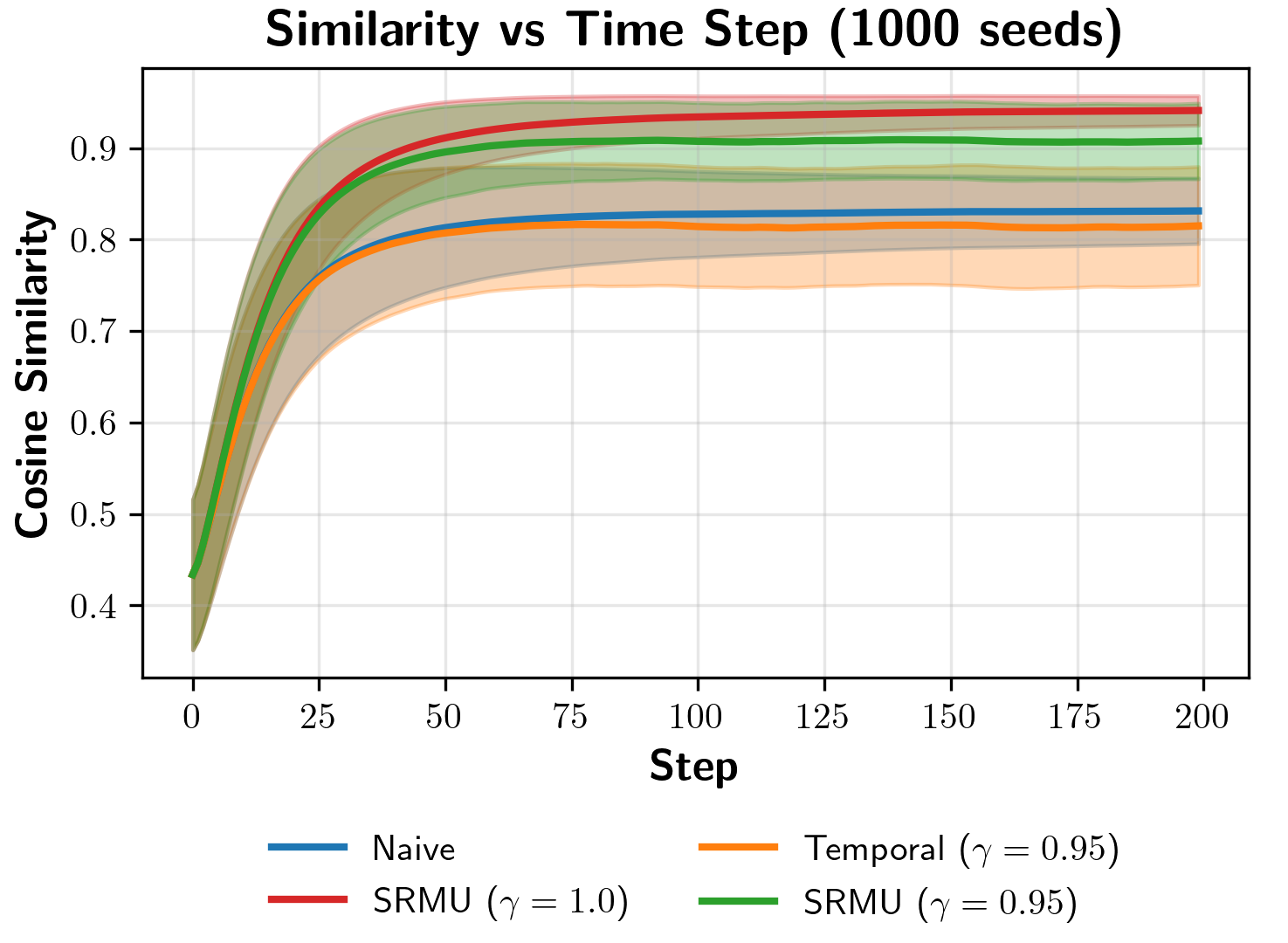}
    \caption{Experiment 1 - Non-Uniform Sampling with 1000 trials.}
    \Description{Experiment 1 - Non-Uniform Sampling with 1000 trials.}
    \label{fig:exp1}
\end{figure}

\vspace{5pt}
\noindent
\textbf{Non-Uniform Sampling }
The non-uniform sampling experiment isolates the effect of sampling imbalance without temporal state changes.
In this configuration, the latent states are fixed over time ($p_\text{drift} = 0$, $p_\text{jump}=0$).
The sampling generator partitions device sampling probabilities into frequent, medium, and sparse groups, before normalizing to a probability distribution.
Because the generator guarantees that every device is observed at least once with the remaining steps allocated based on the sampling probabilities, this experiment creates repeated observations for some devices and sparse observations for others.

This experiment tests if the memory can remain representative of the device population when the incoming observation stream is dominated by a set of frequently observed devices.
A naive additive memory is expected to over-represent frequent devices because repeated observations are always included.
A temporal baseline may partially limit magnitude growth through temporal decay, but decay alone doesn't explicitly account for whether observations contain new information. The SRMU is intended to address this issue by reducing the influence of redundant information and emphasizing new information.

As shown in Table~\ref{table:exp-results} and Figure~\ref{fig:exp1}, clear differences emerge in how each memory model responds to sampling imbalance.
The naive additive model achieves moderate alignment with the ground-truth memory (cosine similarity = 0.832).
However, this performance is accompanied by substantial memory growth ($\|M\| \approx 1650$).
This indicates that accuracy is driven by repeated reinforcement of frequently observed devices rather than efficient representation.

The temporal decay baseline ($\gamma = 0.95$) significantly reduces memory magnitude ($\|M\| \approx 168.93$).
However, it also slightly reduces alignment (cosine similarity = 0.817).
This suggests that uniform decay suppresses both redundant and informative contributions.

In contrast, the SRMU produces higher alignment with the latent state while maintaining more controlled memory growth.
Without decay ($\gamma = 1.0$), SRMU achieves the highest cosine similarity (0.942).
At the same time, it reduces memory magnitude by more than 50\% compared to the naive model ($\|M\| \approx 726.40$).
This shows that suppressing redundant updates alone improves representational efficiency.
When combined with temporal decay ($\gamma = 0.95$), the SRMU further stabilizes the memory ($\|M\| \approx 75.96$).
It maintains strong alignment (cosine similarity = 0.909) and outperforms the temporal baseline in both metrics.

These results indicate that sampling imbalance primarily introduces redundancy rather than new information. 
Accounting for this redundancy during updates leads to more compact and accurate memory representations.
By conditioning updates on similarity, SRMU avoids over-reinforcing frequently observed devices while preserving information about sparsely observed devices.
This results in improved representational fidelity under non-uniform sampling.

\begin{figure}
    \centering
    \includegraphics[width=0.9\linewidth]{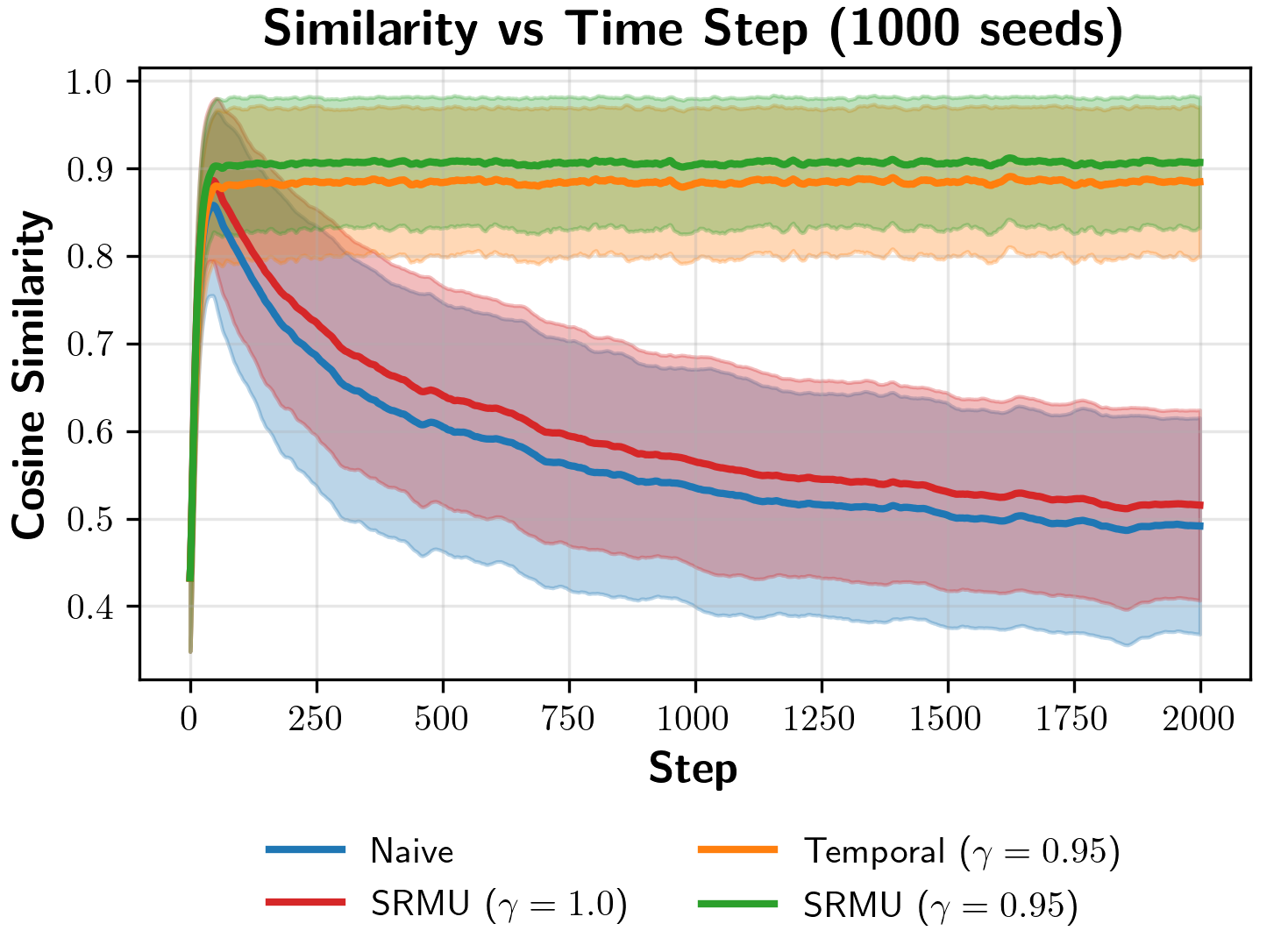}
    \caption{Experiment 2 - Non-Stationary Temporal Dynamics with 1000 trials.}
    \Description{Experiment 2 - Non-Stationary Temporal Dynamics with 1000 trials.}
    \label{fig:exp2}
\end{figure}

\vspace{5pt}
\noindent
\textbf{Non-Stationary Temporal Dynamics }
The non-stationary temporal dynamics experiment isolates the effect of non-stationary latent state dynamics.
Therefore, devices are uniformly sampled, but each device may have local drift ($p_\text{drift}=0.01$, $p_\text{jump} = 0$).
The challenge in this experiment is therefore not sampling imbalance but maintaining a memory that remains consistent with an evolving system.

This experiment is important because the ground-truth memory is reconstructed from the current state of all devices at every timestep, while the memory models only receive one device update per step.
Therefore, a memory that retains all past observations may diverge from the true system state if it cannot appropriately discount outdated information.
The naive baseline has no mechanism to remove stale content, so outdated information accumulates indefinitely.
The temporal baseline introduces global decay, which suppresses obsolete observations and improves adaptation to gradual changes, but it may also attenuate information for rarely sampled devices.
The SRMU is intended to operate between these extremes by combining temporal decay with a relevance-based update mechanism.

As shown in Table~\ref{table:exp-results} and Figure~\ref{fig:exp2}, clear differences emerge in how each memory model responds to non-stationary dynamics.
The naive additive model achieves low alignment with the ground-truth memory (cosine similarity = 0.492). 
This is accompanied by large memory growth ($\|M\| \approx 8115.55$).
This indicates that outdated information accumulates over time, preventing the memory from adapting to evolving system states.

The temporal decay baseline ($\gamma = 0.95$) significantly improves alignment (cosine similarity = 0.884) while maintaining a low memory magnitude ($\|M\| \approx 136.65$).
This shows that temporal decay is effective at removing stale information and enabling adaptation to gradual state changes.
However, this improvement comes from uniformly attenuating all past information, regardless of whether it remains relevant.

In contrast, the SRMU provides a more balanced response to non-stationary dynamics.
Without decay ($\gamma = 1.0$), SRMU slightly improves alignment over the naive model (cosine similarity = 0.515) while reducing memory magnitude ($\|M\| \approx 5665.24$).
This suggests that relevance gating alone can partially limit the accumulation of stale information, but is insufficient to track evolving dynamics.

When combined with temporal decay ($\gamma = 0.95$), SRMU achieves the highest overall performance.
It maintains strong alignment with the latent state (cosine similarity = 0.907) while maintaining a low memory magnitude ($\|M\| \approx 79.06$).
This outperforms the temporal baseline in both alignment and representational efficiency.

These results indicate that non-stationary environments require memory decay and relevance updating. 
Temporal decay enables adaptation but does not distinguish between relevant and outdated information. 
By combining temporal decay with relevance gating, SRMU selectively incorporates new observations while suppressing stale or redundant information.
This results in a memory that remains stable and responsive under evolving system dynamics.

\begin{figure}
    \centering
    \includegraphics[width=0.9\linewidth]{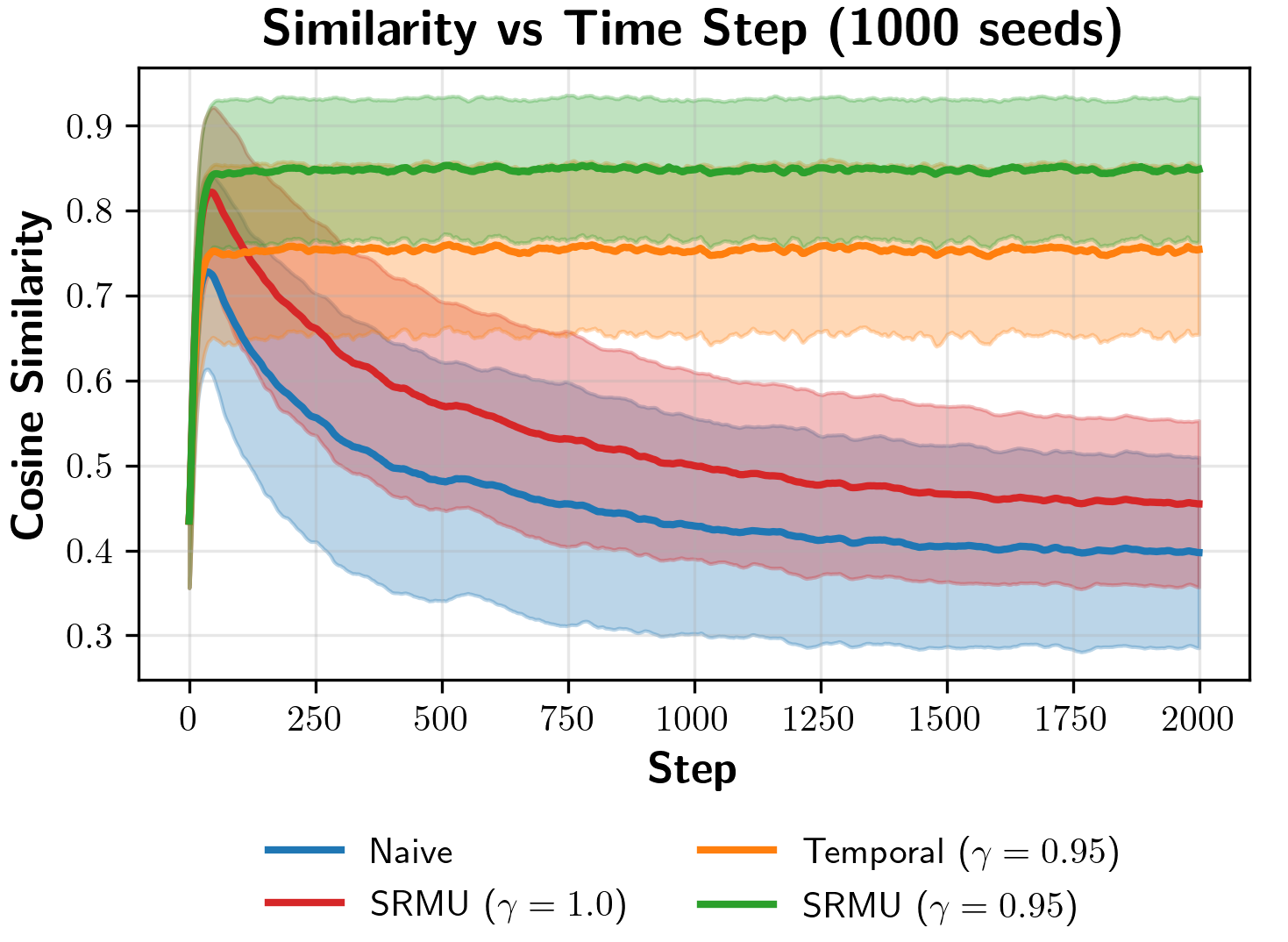}
    \caption{Experiment 3 - Combination with 1000 trials.}
    \Description{Experiment 3 - Combination with 1000 trials.}
    \label{fig:exp3}
\end{figure}

\vspace{5pt}
\noindent
\textbf{Combination Experiment}
The combined experiment evaluates the memory models under a more realistic setting that includes sampling imbalance, non-stationary temporal dynamics, and increased noise.
In this experiment, devices are sampled non-uniformly, while the latent state drifts with probability $p_\text{drift}=0.01$ and jumps with probability $p_\text{jump}=0.001$.

This experiment is the most challenging because the memory models must simultaneously avoid over-reinforcing frequently sampled devices, preserve useful information about infrequently sampled devices, and remain responsive to state changes.
These demands require a trade-off between stability and adaptability.
A memory that is too persistent will retain outdated information, while a memory that forgets too aggressively may fail to preserve the state of rarely sampled devices.

As shown in Table~\ref{table:exp-results} and Figure~\ref{fig:exp3}, the combined experiment produces the clearest separation between the memory models.
The naive model achieves the lowest alignment with the ground-truth memory (cosine similarity = 0.397). It also exhibits the largest memory growth ($\|M\| \approx 9337.07$).
This indicates that the memory accumulates redundant and outdated information, causing severe interference with sampling imbalance and nonstationary dynamics.

The SRMU without decay ($\gamma = 1.0$) provides a modest improvement over the naive baseline.
It increases alignment to 0.455 while reducing memory magnitude to $\|M\| \approx 5670.31$.
This suggests that relevance gating alone can partially suppress redundant updates, but cannot fully compensate for the persistence of stale information in a non-stationary environment.

The temporal decay baseline ($\gamma = 0.95$) improves performance in this experiment.
It reaches a cosine similarity of 0.755 while maintaining a low memory magnitude ($\|M\| \approx 159.26$).
This shows that temporal decay is necessary when the latent system evolves over time.

When combined with temporal decay ($\gamma = 0.95$), the SRMU achieves the best overall result.
It provides the highest cosine similarity (0.850) while also producing the smallest memory magnitude ($\|M\| \approx 74.01$).
This outperforms the temporal baseline in both alignment and representational efficiency.

These results show that the combined experiment requires temporal decay and relevance-gating.
Temporal decay improves adaptation, but it does not account for the information content of new observations.
The SRMU addresses both challenges simultaneously.
By combining decay with relevance gating, the SRMU limits over-reinforcement of frequently observed devices while remaining responsive to state changes.
This produces a more compact and representative memory in environments with non-uniform sampling and non-stationary temporal dynamics.
\section{Conclusion}

Sequential associative memories (SAMs) are difficult to maintain in real-world streaming environments.
Observations arrive sparsely and with nonuniform sampling, while the underlying state may evolve over time.
These conditions create challenges for memory formation: redundant observations can dominate the observation stream, and outdated information may persist after the environment changes.
Vector Symbolic Architectures (VSAs) provide a compact and compositional framework for SAMs.
However, most VSA memory systems rely on simple additive updates that treat all observations as equally informative.
As a result, frequent observations can saturate the representation, while outdated information is not attenuated in non-stationary environments.
This work introduced the Sequential Relevance Memory Unit (SRMU), a domain-agnostic update mechanism for VSA-based SAMs.
The SRMU combines temporal decay with relevance gating to regulate how new observations are incorporated.
This suppresses redundant updates while maintaining responsiveness to new information.
We evaluated the SRMU on streaming state-tracking tasks designed to isolate key challenges in SAMs, including non-uniform sampling, non-stationary dynamics, and their combination.
Across these experiments, the SRMU produced more stable memory growth than the naive additive and temporal decay memory baselines.
These results demonstrate that regulating updates based on their information content improves the robustness of SAMs without relying on domain-specific preprocessing or cleanup mechanisms.
Future work will evaluate the SRMU in larger-scale robotic and distributed sensing systems~\cite{snyder2025generalizable, snyder2026brain}.
Additional research directions include formalizing the capacity and interference properties of relevance-gated updates, and exploring efficient implementations on neuromorphic platforms.

\begin{acks}
We acknowledge the technical and financial support of the
Automotive Research Center (ARC) in accordance with Cooperative Agreement W56HZV-19-2-0001 U.S. Army DEVCOM
Ground Vehicle Systems Center (GVSC) Warren, MI.
\end{acks}

\bibliographystyle{ACM-Reference-Format}
\bibliography{sample-base}










\end{document}